\documentclass[]{article}

\usepackage{graphicx}
\graphicspath{ {./} }
\usepackage[round]{natbib}

\usepackage[vlined,ruled,linesnumbered]{algorithm2e}

\begin{document}

% If your paper is accepted and the title of your paper is very long,
% the style will print as headings an error message. Use the following
% command to supply a shorter title of your paper so that it can be
% used as headings.
%
%\runningtitle{I use this title instead because the last one was very long}

% If your paper is accepted and the number of authors is large, the
% style will print as headings an error message. Use the following
% command to supply a shorter version of the authors names so that
% they can be used as headings (for example, use only the surnames)
%
%\runningauthor{Surname 1, Surname 2, Surname 3, ...., Surname n}

\title{Growing Representation Learning}
\author{Ryan King, Bobak Mortazavi }

\maketitle

\begin{abstract}
  Machine learning continues to grow in popularity due to its ability to learn increasingly complex tasks. However, for many supervised models, the shift in a data distribution or the appearance of a new event can result in a severe decrease in model performance. Retraining a model from scratch with updated data can be resource intensive or impossible depending on the constraints placed on an organization or system. Continual learning methods attempt to adapt models to new classes instead of retraining. However, many of these methods do not have a detection method for new classes or make assumptions about the distribution of classes. In this paper, we develop an attention based Gaussian Mixture, called GMAT, that learns interpretable representations of data with or without labels. We incorporate this method with existing Neural Architecture Search techniques to develop an algorithm for detection new events for an optimal number of representations through an iterative process of training a growing. We show that our method is capable learning new representations of data without labels or assumptions about the distributions of labels. We additionally develop a method that allows our model to utilize labels to more accurately develop representations. Lastly, we show that our method can avoid catastrophic forgetting by replaying samples from learned representations.
\end{abstract}

\section{Introduction}

Machine learning algorithms are powerful tools for a variety of data tasks including classification, compression, denoising, and generation. However, the performance on many state-of-the-art methods \cite{} is a result of fixed models and static datasets. When these models are deployed, they are susceptible to shifts in data distributions \cite{QuioneroCandela2009DatasetSI,DBLP:journals/corr/abs-2107-13964} and perform poorly on new events or classes. In these situations, a new model could be trained with the addition of the new data. However, this process becomes more resource intensive every time a new model needs to be trained. In more constrained situations, old data may not be available due to storage issues or privacy. 

In the real world, data changes either with the appearance of new events or shifts in data distribution. Continual Learning (CL) attempts to adapt models to changes in data without forgetting previously learned information \cite{MCCLOSKEY1989109}. 

Many CL methods exists for adapting models to new events \cite{}. In these methods, models are trained with a subsample of the classes that are available from each dataset. Then a new label is added, and the model is trained with the new data. If a user is aware of a new event, this can be done. However, these methods do not include mechanism for detecting when new events have occurred. A combination of clustering and cluster scoring, such as silhouette score, could be used to identify new events but this requires access to old data which may not be available. Dynamic Gaussian Mixture Models (GMM) have be proposed \cite{DBLP:journals/corr/abs-1808-06508} that form new gaussian mixtures based on a splitting indicator. In the unsupervised setting, these GMMs need to make assumptions about the number of data representations in a data set and often constrain them to be uniform.

In this work, we propose a continual learning algorithm that gradually grows as new representations of data appear which works with or without labels. We accomplish this by first developing an attention mechanism that learns to reconstruct its input data from a set of learned representations of the data we call prototypes. We utilize the weights generated by our attention mechanism to provide interpretability and match prototypes to labels when provided. We then show a method of determining the optimal number of prototypes through an iterative process of training and splitting. During the splitting, we create a network morphism \cite{DBLP:journals/corr/WeiWRC16} of our attention mechanism that can be used to determine the potential improvement from a split and the optimal prototype to be split. The contributions of this paper can be summarized as follows:
\begin{itemize}
\item A novel attention mechanism for learning data representations that we call Gaussian Mixture ATtention (GMAT).
\item A continuous learning scheme that splits our model to an optimal number of representations through NAS.
\item A soft label matching method that allows us to pair learned representations with user inputs.
\end{itemize}

\section{Related Work}

\subsection{Continual Learning}
Data in the real world is always changing either through changes in data distribution \cite{QuioneroCandela2009DatasetSI,DBLP:journals/corr/abs-2107-13964} or through the appearance of new tasks. For machine learning models, this raises the question of what the best policy is for changing models when new data appears. A simple solution would be to retrain the model from scratch with a combination of the data from the old and new tasks. However, this requires organizations to constantly increase their data storage capabilities which may not be feasible. An online embedded model, such as in robotics \cite{DBLP:journals/corr/abs-1907-00182}, a system may not have the ability to store any information. Additional privacy constraints may exist that require data to be destroyed a certain time after its creation \cite{HIPAA}. Continual Learning (CL) attempts to provide a solution to these constraints by updating a model to new information.
    
In CL, as a model is adapted to each new task, old representations have to be remembered without old data. The loss of old representations or previously learned information is called Catastrophic Forgetting \cite{MCCLOSKEY1989109}. One method of combating catastrophic forgetting is called generative replay \cite{DBLP:journals/corr/ShinLKK17,DBLP:journals/corr/abs-1808-06508}. When a adapt a model to new data, generative replay creates represents of old data based on previously learned representations of the data. In \cite{DBLP:journals/corr/ShinLKK17}

\subsection{Gaussian Mixtures}
One way of representing data different classes of data is with a Gaussian Mixture (GM) \cite{reynolds2009gaussian} where the entire data is represented by multiple Gaussian Distributions. Gaussian Mixture Variational Autoencoders have been developed \cite{DBLP:journals/corr/DilokthanakulMG16,DBLP:journals/corr/abs-1910-14481} to learn the parameters of each Gaussian Distributions without labels. However, these methods assume a uniform distribution of labels. In real world data, the actual distribution may be difficult or impossible to determine. In some domains, it's possible that minority classes are more important to classify than the majority classes. An example of this would be in the medical domain where the detection of uncommon health issues need to be detected.

\subsection{Progressive Expansion}
One method of dealing with new tasks is to dynamically expand a model so that new parameters can learn each specific task. In \cite{pmlr-v22-zhou12b}, labeled data is used to train a discriminator and generator. Then a collection of hard examples is collected for data points who's objective function was greater than a predetermined threshold. The model is then expanded based on the number of hard examples collected. In the updated model, new parameters are updated using the hard examples while old parameters are fixed. They additionally, provide a merging method to constrain the size of the network. 

A similar method is used in \cite{DBLP:journals/corr/RusuRDSKKPH16} where the network is updated if the objective function from a new task is higher than some threshold. When the objective function is too high, $k$ units are added to each layer of the network and trained with a sparse regularization to control the growth of the network.

In \cite{DBLP:journals/corr/DraelosMLVCJA16}, an AutoEncoder (AE) is trained using a layerwise reconstruction error. When the network is trained on a new task, the reconstruction error provides an indicator for network growth. When the reconstruction error is large enough, new nodes are added to each level of the AE to allow it to represent the new data. In this method and in \cite{DBLP:journals/corr/RusuRDSKKPH16}, additional nodes may become entangled making it difficult to develop any type of interpretability for each new task.

Neural Architecture Search (NAS) is a field of Automated Machine Learning (AutoML) that attempts to automatically find the optimal neural architecture for a given dataset. Recent methods in NAS \cite{DBLP:journals/corr/abs-1910-02366,DBLP:journals/corr/abs-1910-03103} have shown that many local minimums are high dimensional saddle points. They escape these saddle points by creating a network morphic model \cite{DBLP:journals/corr/WeiWRC16} where potential splitting parameters are represented by multiple parameters while producing the same output for every input. With this representation of their model, they are able to calculate a splitting indicator which can be used to determine the strength of a split at each candidate position and the direction of a split. They show that this method can be coupled with an interpretable machine learning method \cite{DBLP:journals/corr/DraelosMLVCJA16} to find the optimal number of representations in a labeled dataset.

%%%
\section{Methods}

In this section, we break down the components of our method. Specifically, we describe a set of parameters called prototypes and how we train them to represent data. We then describe a hierarchical method for represent our prototypes. Finally, we describe the method we use to grow the number of prototypes.

\subsection{Prototype}
Given some input data $\mathbf{x} \in \mathbf{R}^{N \times L}$, where  we would like to learn how to best represent the data. We choose to model our data as a Mixture of Gaussian \cite{reynolds2009gaussian} defined by: 

$$p(x |\lambda) = \sum_{i=1}^M w_i g(x|\mu_i, \sigma_i)$$  

Where $M$ is the number of prototype in our mixture, $w_i$ is the weight of each Gaussian distribution, $\mu_i$ is the mean, $\sigma_i$ is the standard deviation and $\lambda$ is the collection of $w$, $\mu$, and $\sigma$ associated with each individual Gaussian.  Weights are constrained so that $\sum w_i = 1$. This makes the probability of $x \in \mathbf{x}$ the weighted average of Gaussian Distributions in the mixture. In our implementation, we consider each Gaussian Distribution a prototype.

Traditionally, the values of $\mu$ and $\sigma$ are the output of a trained linear layer. In each of our prototypes, we choose to represent $\mu$ and $\sigma$ as trainable parameters. To train these parameters to represent the data, we attempt to reconstruct the data with a combination of each prototype. We accomplish this by first taking the Mahaloanobis Distance \cite{mahalanobis1936generalized} between $X$ and $P$ defined as:

\begin{equation}
    D(x) = \sqrt{(\mathbf{x} - \mathbf{\mu_i})^{\mathbf{T}}\Sigma_i^{-1}(\mathbf{x} - \mathbf{\mu_i})}
\end{equation}

With these distances, we consider data points that are close to the mean of a particular prototype to be highly represented by that prototype. With this intuition, we use the softmin equation to determine what percentage of each prototype, $\alpha$, should be used to reconstruct the data point:

\begin{equation}
    w_i(\mathbf{x}) = \frac{e^{-D(x_i)}}{\sum_{j=1}^L{e^{-D(x_j)}}}
\end{equation}

Once we have derived the weights, we can take the weighted average, $\mathbf{z}_i$, of the prototypes to reconstruct the data point.

\begin{equation} \label{eq:attn}
    \mathbf{z}_i(\mathbf{x}) = w_i(\mathbf{x}) \cdot P
\end{equation}

\begin{figure}[h!]
  \includegraphics[scale=0.6]{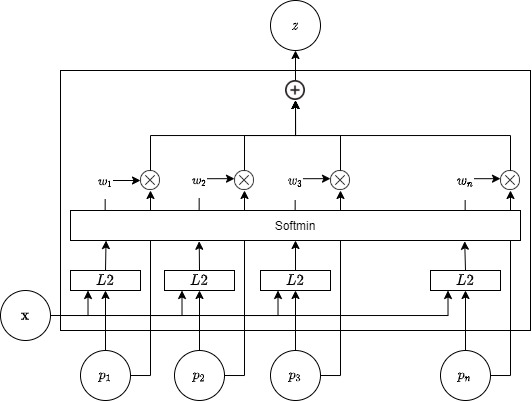}
  \caption{A visual illustration of our GMAT mechanism. We calculate the weights by first measuring the distance between a data point $\mathbf{x}$ and each prototype distribution. The softmin is then appplied to the distances to determine the weight of each distribution in the Gaussian Mixture. The weighted sum of the prototypes is used to reconstruct the value $\mathbf{x}$}
  \label{fig:attnblock}
\end{figure}

This mechanism, which we call GMAT, acts similar to soft attention where a importance weight is applied to each feature. However, in our method we take the softmin of the L2-norm to enable our attention mechanism create weight based on a prototypes closeness to a particular distributions mean. A graphic representation of our attention block is in Figure \ref{fig:attnblock}

Given this setup, we'd like to reconstruct a data point based on some combination of the prototypes. We measure the distance between $x$ and each prototype as the Mean Square Error given as:

$$\mathcal{L}_{recon}(\mathbf{x}, \mathbf{z}) = \frac{1}{N} \sum_{i = 1}^{N}||x_i - z_i||^2_2$$

Where $\mu_i$ is the mean of each prototype and $\Sigma_i$ is the covariance matrix generated by diagonalizing $\sigma^2_i$. Since prototype that are close to a point provide a better representation of that point, we want them to be weighted more heavily. With this intuition, we use the softmin of the distances to generate the weights for each prototype. We use the reparameterization trick \cite{kingma2014autoencoding} by randomly sample a point from each of our prototypes. Weights from our softmin are applied to the randomly sampled points.

We make further use of the importance weights in our modified KL Divergence term. We start by calculating the weighted mean, $\mu_i$ and variance, $\sigma^2$, with $w_i(\mathbf{x})$ for each prototype:

\begin{equation} \label{eq:w_mean}
    \hat{\mu}_i = \frac{\sum_{j=1}^N w(\mathbf{x})_{i,j} \mathbf{x}}{\sum_{j=1}^N w(\mathbf{x})_{i,j}}
\end{equation}

\begin{equation} \label{eq:w_var}
    \hat{\sigma}_i^2 = \frac{\sum_{j=1}^N w(\mathbf{x})_{i,j} (\mathbf{x}_i - \mu_i(\mathbf{x}))^2}{\sum_{j=1}^N w(\mathbf{x})_{i,j}}
\end{equation}

We use equations \ref{eq:w_mean} and \ref{eq:w_var} to create a target Normal Distribution, $P_i$, of $x$ for each prototype:

\begin{equation} \label{eq:w_var}
    \mathbf{P} = \mathcal{N}(\hat{\mu}_i, \hat{\sigma}_i^2)
\end{equation}

We then use this distribution to calculate the KL Divergence between the our prototypes, $\mathbf{Q}_i$, and the weighted distribution of the data.

\begin{equation} \label{eq:w_var}
    \mathbf{KL}(\mathbf{P}_i || \mathbf{Q}_i)
\end{equation}

We additionally add two terms, equations \ref{eq:R1} and \ref{eq:R2}, to our loss function that increase the interpretability and certainty of our model. 

\begin{equation} \label{eq:R1}
    \mathcal{L}_{R_1} = -\frac{1}{N} \sum_{i=1}^{N} \mathbf{y}_i log(\mathbf{Z}_{[i,:]})
\end{equation}

\begin{equation} \label{eq:R2}
    \mathcal{L}_{R_2} = -\frac{1}{M} \sum_{j=1}^{M} \mathbf{y}_j log(\mathbf{Z}_{[:,j]})
\end{equation}

Where $z_i \in \mathbf{Z}$, $y_i$ and $y_j$ is a one hot encoding of the batch-wise argmax and prototype-wise argmax of $\mathbf{Z}$. Both of these equations treat the weights from $z_i$ as probabilities. In equation \ref{eq:R1}, we take the Negative Log Likelihood in an attempt to avoid saddle points where the model may want to use a combination of prototypes to reconstruct the output. Equation \ref{eq:R2} increases the certainty that each prototype is used to reconstruct the data.

\subsection{Progressive Splitting}
We have a way of training our model to learn the weights of each prototype. We would now like to learn the optimal number of prototypes in our data. In order to accomplish, we develop an iterative process of training the parameters of our network, testing for potential growth and splitting if necessary. The algorithm for our splitting method can be seen in Algorithm \ref{algo_splitting}. In previous sections we discuss how our GMAT mechanism is trained. This section will be broken down into the steps following training which include the creation of a network morhpic model to test for growth and a splitting indicator for growth.

\begin{algorithm}
\caption{Optimal Splitting}
\SetKwData{Left}{left}
\SetKwData{This}{this}
\SetKwData{Up}{up}
\SetKwFunction{Union}{Union}
\SetKwFunction{FindCompress}{FindCompress}
\SetKwInOut{Input}{input}
\SetKwInOut{Output}{output}
\Input{\{X, Y\}}
initialize $\theta$ \\
S = $\infty$ \\
\While{ $S > \epsilon$}{
    \While{ not converged }{
        train($\theta$, X, Y)
    }
    for 
    $S = \nabla L(X, \theta)$
}
\label{algo_splitting}
\end{algorithm}

\subsubsection{Network Morphism}
In order to test our for potential growth, we create a network morphism \cite{DBLP:journals/corr/WeiWRC16} of our GMAT mechanism by creating a set of potential splitting position we call dummies with the constraint that the new network structure $G$ must perform the same mapping as our original GMAT mechanism $F$. For both our linear and binary tree structure, we create $G$ by copying each prototype in $F$. For our linear structure, we simply copy the value of $\mathbf{\mu}$ for each prototype.

In order to prove that the addition of new prototypes results in an network morphic representation of our prototype layer, we provide the following proof. Given that for $\hat{x} = \{x, x\}, \hat{P} = \{P, P\}$.

\begin{equation}
    \sum_{m \in M} \frac{P_m e^{x_m}}{\sum_{j \in M} e^{x_j}} = \sum_{m \in 2M} \frac{\hat{P}_m e^{\hat{x}_m}}{\sum_{j \in 2M} e^{\hat{x}_j}}\\
\end{equation}

\begin{equation}
    \sum_{m \in 2M} \frac{\hat{P}_m e^{\hat{x}_m}}{\sum_{j \in 2M} e^{\hat{x}_j}}
\end{equation}

Since $\hat{x} = \{x, x\}$, it is equivalent to write the following:

\begin{equation}
    \sum_{m \in 2M} \frac{\hat{P}_m e^{\hat{x}_m}}{2\sum_{j \in M} e^{x_j}}
\end{equation}

The same claim can be made for the outer summation where $\hat{x} = \{x, x\}$ and $\hat{P} = \{P, P\}$ giving the following:

\begin{equation}
    2\sum_{m \in M} \frac{P_m e^{x_m}}{2\sum_{j \in M} e^{x_j}}
\end{equation}

The final reduction gives the following:

\begin{equation}
    \sum_{m \in M} \frac{P_m e^{x_m}}{\sum_{j \in M} e^{x_j}} = \sum_{m \in M} \frac{P_m e^{x_m}}{\sum_{j \in M} e^{x_j}}
\end{equation}

With this isomorphic model representation, we can determine which additional branch would produce the best improvement. We do this by taking the average magnitude of the gradient from a pass through a dataset. By doing this, we 

\subsubsection{Splitting Indicator}
With our network morphic structure, we are now able to represent each prototype as multiple prototypes. We would now like to understand how much each prototype in $G$ can potential help to increase the performance of $F$. Given the models parameters from $G$, $\hat{\theta}$, we calculate the average direction $D$ of the gradient to determine the splitting direction:

$$D(x,\hat{\theta}) = \frac{1}{N} \sum_{i=1}^{N} \nabla L(\mathbf{x}_i, \hat{\theta})$$

We use the magnitude of the gradient of $G$ as an indicator of the strength, $S$, of each split:

$$S(x,\hat{\theta})=||D(x,\theta)||^2_2$$

As prototypes are split to create new representations of the data, the magnitude of the splitting indicator decreases. We use $S$ as an indicator of when to stop split. We choose a threshold parameter $\epsilon$ to indicate when splitting should stop.

\subsection{Model Architecture}
Our GMAT mechanism can be used by itself to detect Gaussian Distributed classes from streams of data. However, we include an encoder $E(x)$ and decoder $D(z)$ so that the dimensionality of the input data can be reduced. We show in the section on Label Matching that the encoder can be used to map inputs to their respective average latent representations with the use of labels. For 

\begin{figure}[h!]
  \includegraphics[scale=0.7]{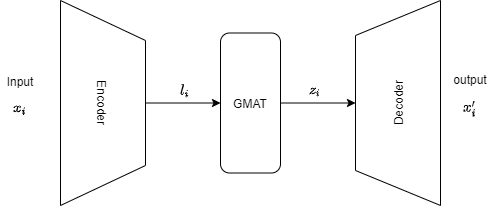}
  \caption{A visual representation of our model architecture. Input data is fed in a decoder which produces a latent representation. The latent representation is used as the input to our GMAT mechanism. GMAT reconstructs the representation for a combination of prototype distributions to produce $z$. This becomes the input for a decoder that reproduces $x$.}
  \label{fig:modelarch}
\end{figure}

A visual representation of our model architecture is shown in Figure \ref{fig:modelarch}

An appropriate encoder and decoder architecture can be chosen based on the input data. In our experiments we use a linear architecture to show how or model does on two simulated datasets. Additionally, we use a convolutional architecture in our experiments on MNIST.

\subsection{Generative Replay}
As our model adapts to new data, we need to remember previously learned representations. Generative Replay methods \cite{DBLP:journals/corr/ShinLKK17,DBLP:journals/corr/abs-1808-06508} have been successful in combating catastrophic forgetting by training on generated representations of old data. In DGR \cite{DBLP:journals/corr/ShinLKK17}, a generator is used to replay old representations of data while a solver is trained to classify. Other adaptations have been developed \cite{DBLP:journals/corr/abs-1808-06508} that use the decoder of a VAE architecture to generate samples from a Gaussian Mixture prior called Mixture Generative Replay MGR. We develop an adaptation of MGR using our prototypes. During training, we add generated into our dataset and treat it as training data. We generate data from our decoder by uniformly sampling each prototype learned from previous tasks. Samples from each prototype are randomly selected based on the $\mu$ and $\sigma$ describing their distribution.'

\subsection{Label Matching}
We provide an unsupervised method for learning an optimal number of representations of data. It's possible that data has labels or labels can be generated as prototypes become interpretable. However, the number of labels that are in a dataset may not match the number of prototypes currently trained. In the opposite situation, we may have a label for each prototype but we want to see if new representations of data exist. Since we can potentially have a different number of prototypes and labels, we would provide a soft labeling mechanism that works with our attention mechanism to match prototypes to labels. 

We need each class of data to be Gaussian Distributed in order for it to be represented by a prototype. We utilize any provided labels to compute the per batch average of each class. We then measure the distance of each point to each average representation in the latent space.

$$d(x, L_{avg}) = ||\mathbf{x} - L_{avg}^i||^2_2$$

We use the softmin function similar to the way prototype importance is determined in the unsupervised setting.

$$\frac{e^{d(x_i)}}{\sum_{j=1}^L e^{d(x_j)}}$$

We treat the output of the softmin function as the probability that each data point, $E(x_i)$, belongs to each latent representation $l_n$. With these probabilities and provided labels, we calculate the classification loss using Cross Entropy. When label matching is being used, the encoder learns to map each $x_i$ into clusters centered around their latent average. 

\section{Experiments}
We create our model and implement our splitting method using PyTorch \cite{NEURIPS2019_9015}. Our model, methods and experiments can be found at github.com. For all of our 

\subsection{Datasets and Evaluation}
MNIST is an image dataset of handwritten digits from 0 to 9. We use this dataset throughout our experiments for supervised, unsupervised and continual learning. We use decode the means of each prototype to get an understanding of the data each distribution is representing. For all of our experiments with MNITS, we use a 4 layer CNN for our encoder and decoder with channel sizes of 32, 32, 32, and 10.

We evaluate each of our methods using the Normalized Mutual Information score on unsupervised tasks:

$$NMI(Y, C) = \frac{2 \times I(Y, C)}{[H(Y) + H(C)]}$$

Where $Y$ are the true labels, $C$ are the predicted labels, $H$ is the entropy, and $I(Y, C)$ is the mutual information between $Y$ and $C$. We also calculate the accuracy for comparison with other supervised methods.

\subsection{Simulated Dataset}
We use two simulated datasets to provide some intuition behind what each part of our model is doing. The first is a mixture of isotropic Gaussian Distributed classes. We train our GMAT mechanism alone on this dataset to show that it is capable of identifying Gaussian Distributed clusters from streams of data. The second dataset is the half moon dataset which we use with our entire model. We use this dataset to show that labels can be used to help our prototypes identify non-Gaussian distributed classes. We generate both of these datasets using the scikit-learn library \cite{scikit-learn}. We use a simple linear Autoencoder with a single hidden layer of size 100 and a latent size of 2 for easy visualization. Details about the data generation and training for each dataset can be found in Appendix... In Figure \ref{fig:sim_data} we show that the the mean of each prototype is centered in each cluster and the confidence intervals are drawn around those points. 

\begin{figure}[h!]
  \includegraphics[scale=0.8]{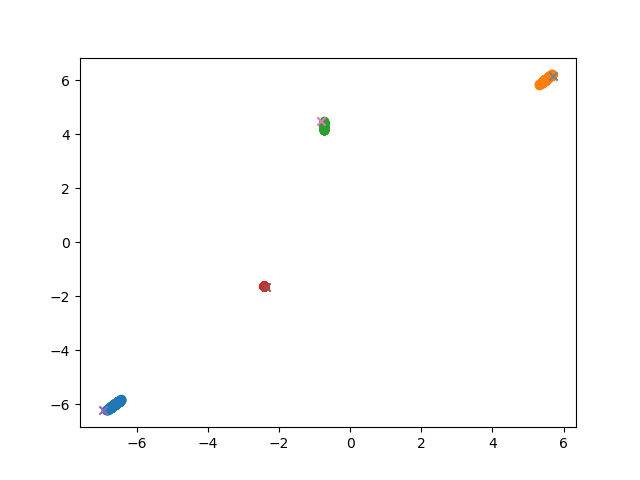}
  \caption{GMAT trained on simulated dataset. In this figure, each color represent a different class of data. The means of our prototype distributions are represented by 'x' which are centered on each class.}
  \label{fig:sim_data}
\end{figure}

\subsection{Optimal Splitting}
We train our model on every class in the MNIST dataset starting with 1 prototype distribution and allowing our network to split until an optimal NMI is reached. We collect the average magnitude per gradient and NMIs at each iteration and plot them in Figure \ref{fig:mags_per_iter}
bar graph with error

\subsection{Supervised Learning}
\begin{figure}[h!]
  \includegraphics[scale=1.25]{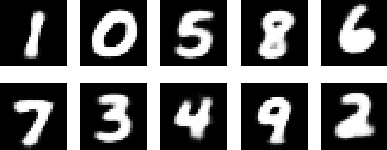}
  \caption{Decoded means of each prototype trained with labels.}
  \label{fig:sup_prot}
\end{figure}

Incremental class learning and incremental task learning \cite{DBLP:journals/corr/abs-1904-07734,DBLP:journals/corr/abs-1810-12488} are popular methods of evaluating a CL methods. Both methods evaluate a models ability to learn new data without forgetting previously learned information by training a model by incrementally training a model on a subset of . In incremental class learning, a model is To evaluate our model, we use the SplitMNIST dataset, where the MNIST dataset is divided into 5 separate datasets by class (0/1, 2/3, 4/5, 6/7, 8/9). We train each iteration for up to 500 epochs with a patience of 50 epochs without improvement in the loss function. We compare our method with other Continuous Learning methods on the same task and report the results in Table \ref{tab:inc_class}. The prototypes resulting from our supervised method are shown in Figure \ref{fig:sup_prot} \footnote{Results are taken from \cite{DBLP:journals/corr/abs-1910-14481}, \cite{DBLP:journals/corr/abs-1810-12488}, and \cite{DBLP:journals/corr/abs-1904-07734}}.

% \begin{table}[]
%     \centering
%     \begin{tabular}{|c|c|}
%     \hline
%     Method                      & Accuracy       \\ \hline
%     EWC                         & 19.80 $\pm$ 0.05 \\ \hline
%     SI                          & 19.67 $\pm$ 0.09 \\ \hline
%     MAS                         & 19.52 $\pm$ 0.29 \\ \hline
%     LwF                         & 24.17 $\pm$ 0.33 \\ \hline
%     GEM                         & 92.20 $\pm$ 0.12 \\ \hline
%     DGR                         & 90.79 $\pm$ 0.41 \\ \hline
%     iCARL                       & 94.57 $\pm$ 0.11 \\ \hline
%     CURL                        & 92.59 $\pm$ 0.66 \\ \hline
%     GMAT                        & 00.00 $\pm$ 0.00  \\ \hline
%     \end{tabular}
%     \caption{Supervised learning on SplitMNIST for incremental class learning.}
%   \label{tab:inc_class}
% \end{table}

\subsection{Unsupervised Clustering}
\begin{figure}[h!]
  \includegraphics[scale=0.75]{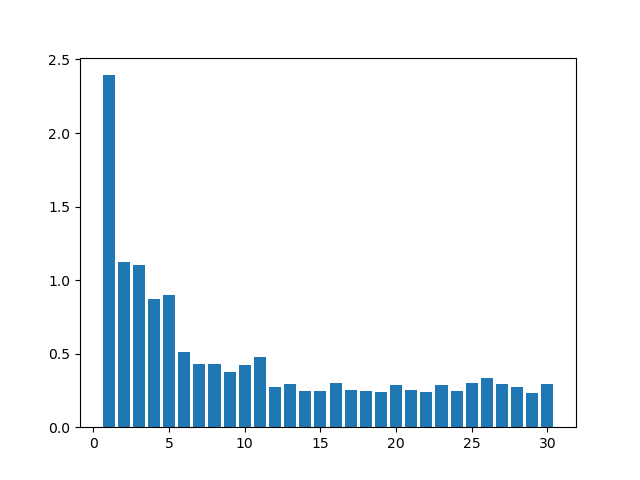}
  \caption{Unsupervised splitting}
  \label{fig:unsup_bar}
\end{figure}

We test our method on it's ability to cluster data without labels by training our model on the MNIST dataset. We initialize our model with a single prototype and allow is to continue splitting for 30 iterations. We collect the splitting strength at each iteration and the NMI. We report the values in \ref{fig:unsup_bar}

We found that as the strength of our splitting indicator decreases, the increase in the NMI between iterations also decreases.

\bibliography{Reference}

\end{document}